\pdfoutput=1

\documentclass[11pt]{article}

\usepackage[]{EMNLP2022}

\usepackage{times}
\usepackage{latexsym}

\usepackage[T1]{fontenc}

\usepackage[utf8]{inputenc}

\usepackage{microtype}

\usepackage{bm}
\usepackage{amsmath}
\usepackage{url}
\usepackage{graphicx}
\usepackage{float} 
\usepackage{subfigure}
\usepackage{multirow}
\usepackage{makecell}
\usepackage{xpinyin}
\usepackage{enumerate}

\usepackage{xcolor}
\usepackage{framed}
\usepackage{color}
\usepackage{bbding}
\usepackage{soul}

\usepackage{colortbl}
\newcolumntype{L}[1]{>{\raggedright\let\newline\\\arraybackslash\hspace{0pt}}m{#1}}
\setul{2pt}{2pt}
\usepackage{booktabs}
\usepackage{amsmath,amsfonts}
\usepackage[utf8]{inputenc}
\usepackage{cleveref}

\usepackage{ulem}
\usepackage{diagbox}
\crefname{section}{§}{§§}
\Crefname{section}{§}{§§}
\usepackage{enumitem}
\setenumerate[1]{itemsep=0pt,partopsep=0pt,parsep=\parskip,topsep=5pt}
\setitemize[1]{itemsep=0pt,partopsep=0pt,parsep=\parskip,topsep=5pt}
\setdescription{itemsep=0pt,partopsep=0pt,parsep=\parskip,topsep=5pt}
\usepackage{arydshln}

\usepackage[linesnumbered,ruled,vlined]{algorithm2e}
\usepackage{algorithmic}

\title{BJTU-WeChat's Systems for the WMT22 Chat Translation Task}

\author{
  Yunlong Liang\textsuperscript{1}\thanks{ \ \ Work was done when Yunlong was interning at Pattern Recognition Center, WeChat AI, Tencent Inc, China.}  ,  
  Fandong Meng\textsuperscript{2}, 
  \textbf{Jinan Xu}\textsuperscript{1}\thanks{ \ \ Jinan Xu is the corresponding author.}  , 
  \textbf{Yufeng Chen}\textsuperscript{1}\ and \textbf{Jie Zhou}\textsuperscript{2}\\
  \textsuperscript{1}Beijing Key Lab of Traffic Data Analysis and Mining, \\Beijing Jiaotong University, Beijing, China \\
  \textsuperscript{2}Pattern Recognition Center, WeChat AI, Tencent Inc, China \\
  \texttt{\{yunlongliang,jaxu,chenyf\}@bjtu.edu.cn} \\
  \texttt{\{fandongmeng,withtomzhou\}@tencent.com} \\
}

\begin{document}
\maketitle
\begin{abstract}
This paper introduces the joint submission of the Beijing Jiaotong University and WeChat AI to the WMT'22 chat translation task for English$\Leftrightarrow$German. Based on the Transformer~\cite{vaswani2017attention}, we apply several effective variants. In our experiments, we utilize the pre-training-then-fine-tuning paradigm. In the first pre-training stage, we employ data filtering and synthetic data generation (i.e., back-translation, forward-translation, and knowledge distillation). In the second fine-tuning stage, we investigate speaker-aware in-domain data generation, speaker adaptation, prompt-based context modeling, target denoising fine-tuning~\cite{meng-etal-2020-wechat}, and boosted self-COMET-based model ensemble. Our systems achieve 0.810 and 0.946 COMET~\cite{rei-etal-2020-comet} scores\footnote{The COMET is the official automatic evaluation metric.} on English$\rightarrow$German and German$\rightarrow$English, respectively. The COMET scores of English$\rightarrow$German and German$\rightarrow$English are the highest among all submissions.
\end{abstract}

\section{Introduction}
We participate in the WMT 2022 shared task on chat translation in two language directions, English$\rightarrow$German and German$\rightarrow$English. In this year's chat translation task, we apply the two-stage training strategy. In the first stage, we investigate model architecture and data augmentation. In the second stage, we mainly focus on exploiting speaker-aware in-domain data augmentation, speaker adaptation, prompt-based context modeling, target denoising fine-tuning~\cite{meng-etal-2020-wechat}, and model ensemble strategies. This task aims to build machine translation systems to translate conversational text and thus supports fluent communication between an agent speaking in English and a customer speaking in a different language (e.g., German), which is different from the first pre-training stage~\cite{farajian-etal-2020-findings,liang-etal-2021-modeling,liang2022msctd,liu2021recent,gain2021not,gain2022low,buschbeck2022hi}. Therefore, we mainly pay attention to the second fine-tuning stage.

In the first pre-training stage, we follow previous work~\cite{meng-etal-2020-wechat,zeng-etal-2021-wechat,meng2019dtmt,yanetal2020multi} and utilize several effective Transformer variants. Specifically, we combine the Multi-Head-Attention~\cite{vaswani2017attention}, Average Attention Transformer~\cite{zhang-etal-2018-accelerating}, and Talking-Heads Attention~\cite{https://doi.org/10.48550/arxiv.2003.02436}, which have shown significant model performance and diversity. For data augmentation, we employ the back-translation method to use the target-side monolingual data and apply the forward-translation to leverage the source-side monolingual data. To fully utilize the source-side of bilingual data, we use the sequence-level knowledge distillation method~\cite{kim-rush-2016-sequence}.


In the second fine-tuning stage, for speaker-aware in-domain data augmentation, based on the BConTrasT~\cite{farajian-etal-2020-findings} dataset of the WMT20 chat translation task, we firstly adapt our pre-trained model to each speaker by using the speaker tag as a pseudo token and then apply it to the Taskmaster-1~\cite{byrne-etal-2019-taskmaster} corpus to generate the speaker-aware in-domain data. For speaker adaptation, we follow previous work~\cite{moghe-hardmeier-bawden:2020:WMT} to prepend the corresponding speaker tag to each utterance on both the source and the target side to get a speaker-aware dataset. For prompt-based context modeling, we exploit the prompt learning to incorporate the bilingual context and then apply the target denoising fine-tuning method~\cite{meng-etal-2020-wechat} to train our model. For the model ensemble, inspired by~\citet{zeng-etal-2021-wechat}, we select high-potential candidate models from two aspects, namely model performance (COMET scores) and model diversity (Self-COMET scores among all candidate models). Based on this, we design a search algorithm to gradually select the current best model of the model candidate pool for the final model ensemble.


\section{Model Architectures}

In this section, we describe the model architectures we used in two translation directions, where we mainly follow the previous state-of-the-art models~\cite{zeng-etal-2021-wechat}. We also refer readers to read the paper for details.

\subsection{Model Configurations}
Given the strong capacity of deeper and wider architectures, we use them in our experiments. Specifically, following~\citet{zeng-etal-2021-wechat}, we use 20-layer encoders for deeper models and set the hidden size to 1024 for all models. We set the decoder depth to 10. For the wider ones, we adopt 12 encoder layers, 2048 for hidden size, and 8192 to 15000 for filter sizes.

\subsection{Transformer Variants}

\paragraph{Average Attention Transformer.}

Following~\citet{zeng-etal-2021-wechat}, the average attention transformer~\cite{zhang-etal-2018-accelerating} are employed to add model diversity. In the AAN, the context representation $g_i$ for each input embedding is calculated as follows:
\begin{equation}
\label{eq}
    g_i = \mathrm{FFN}(\frac{1}{i}\sum_{k=1}^{t}y_k), \nonumber
\end{equation}
where $y_k$ is the input embedding for step $k$ and $t$ is the current time step. $\mathrm{FFN}$ denotes the position-wise feed-forward network~\cite{vaswani2017attention}. 

\paragraph{Talking Heads Attention.}
Similarly, talking-heads attention~\cite{https://doi.org/10.48550/arxiv.2003.02436} also performs well in~\citet{zeng-etal-2021-wechat}, which can transform the attention-logits and the attention scores and thus allow information interaction among attention heads by adding two linear projection layers $W_l$ and $W_a$:
\begin{equation}
\label{eq}
    \mathrm{Attention}(Q, K, V) = \mathrm{softmax}(\frac{QK^T}{\sqrt{k}}W_l)W_aV. \nonumber
\end{equation}

\section{System Overview}
In this section, we describe our system used in the WMT 2022 chat translation shared task, which includes two parts, namely general pre-training and in-domain fine-tuning. The pre-training part includes data filtering and synthetic data generation. The in-domain fine-tuning consists of speaker-aware in-domain data generation, speaker adaptation, prompt-based context modeling, the target denoising fine-tuning~\cite{meng-etal-2020-wechat}, and boosted Self-COMET-based model ensemble.

\subsection{General Pre-training}

\subsubsection{Data Filtering}
\label{sec:data_filter}
We filter the bilingual training corpus (including synthetic parallel data) with the following rules~\cite{zeng-etal-2021-wechat}: 1) Normalize punctuation; 2) Remove the sentence whose length is more than 100 words or a single word that exceeds 40 characters; 3) Filter out the duplicated sentence pairs; 4) Delete the sentence whose word ratio between the source and the target words exceeds 1:4 or 4:1. 

\subsubsection{Synthetic Data Generation}
For data augmentation, we obtain the general domain synthetic data via back-translation, forward-translation, and knowledge distillation.

\paragraph{Tagged Back-Translation.}
Previous work has shown that different methods of generating pseudo corpus have a different influence on translation performance~\cite{edunov-etal-2018-understanding,hoang2018iterative,zeng-etal-2021-wechat}. Following them, we attempt two generating strategies: 1) Beam Search: produce translation by beam search (beam size = 5). 2) Sampling Top-k: Select a word randomly from top-k (k = 15) words when inference.

\paragraph{Forward-Translation.}
We then ensemble models to forward-translate the monolingual data of the source language to further enhance model performance. We obtain a stable improvement in both directions, which is consistent with previous work~\cite{zeng-etal-2021-wechat}. 

\paragraph{Knowledge Distillation.}
Knowledge Distillation aims to transfer knowledge from the teacher model to student models, which has shown effective for NMT~\cite{kim-rush-2016-sequence,wang2021selective,zeng-etal-2021-wechat}. Specifically, we first use the teacher model to generate synthetic corpus in the forward direction (i.e., En$\to$De). Then, we train our student models with the generated corpus. 

Note that we prefix all the synthetic sentences by appending a pseudo tag \texttt{<BT>} when jointly training with genuine data.

\subsection{In-domain Fine-tuning}

\subsubsection{Speaker-aware In-domain Data Generation}
\label{speaker-aware}
Inspired by~\citet{moghe-hardmeier-bawden:2020:WMT}, we prepend the corresponding speaker tag (the \texttt{<agent>} or the \texttt{<customer>}) to each utterance on both the source and the target side to get a speaker-aware dataset based on the BConTrasT dataset of the WMT20 chat translation task~\cite{farajian-etal-2020-findings}. Secondly, we adapt our pre-trained model to each speaker on the speaker-aware dataset. Then, we apply the adapted model to the monolingual Taskmaster-1~\cite{byrne-etal-2019-taskmaster} corpus, which is the original source of BConTrasT~\cite{farajian-etal-2020-findings}, to generate the speaker-aware in-domain data. 

\subsubsection{Speaker Adaptation}
As a special characteristic of chat translation, distinguishing between the two speaker roles plays an important role as they both form the complete dialogue. And modeling the speaker characteristic has been demonstrated effective in previous work~\cite{moghe-hardmeier-bawden:2020:WMT,liang-etal-2021-towards,liang-etal-2022-scheduled,liang2021infusing,liang2022emotional}. Therefore, our data used in the fine-tuning has a corresponding speaker tag (the \texttt{<agent>} or the \texttt{<customer>}) appended in the first token of each utterance. 

\subsubsection{Prompt-based Context Modeling}
Previous studies~\cite{wang-EtAl:2020:WMT1,moghe-hardmeier-bawden:2020:WMT} have shown that the multi-encoder framework cannot improve the model performance after using the context in the chat translation task, while a unified model~\cite{ma-etal-2020-simple,liang-etal-2021-towards} can. Therefore, we also investigate incorporating the context in the unified model with prompt learning (without modifying the model architecture). Specifically, we add two preceding bilingual contexts at the tail of each utterance with an indicator \texttt{<context  begins>}, where we also use a special tag \texttt{<SEP>} to separate different utterances of the bilingual context. In this way, our model with context modeling can achieve a better COMET.

\subsubsection{Target Denoising Fine-tuning}
To bridge the exposure bias~\cite{ranzato2015exposurebias}, we add noisy perturbations into decoder inputs when fine-tuning. Therefore, the model becomes more robust to prediction errors by target denoising fine-tuning~\cite{zhang-etal-2019-bridging,meng-etal-2020-wechat}.
Specifically, the fine-tuning data generator chooses 30\% of utterance pairs (Note that we do not include the indicator word and the bilingual context) to add noise and keeps the remaining 70\% of sentence pairs unchanged. For a chosen pair, we keep the source sentence untouched and replace the $i$-th token of the target sentence with (I) a random token of the current target sentence in 15\% probability and (II) the unchanged $i$-th in 85\% probability.

\begin{algorithm}[t!]
	\renewcommand{\algorithmicrequire}{\textbf{Input:}}
	\renewcommand{\algorithmicensure}{\textbf{Output:}}
	\caption{Boosted Self-COMET-based Ensemble (BSCE)}
	\label{alg:1}
	\begin{algorithmic}[0]
        \REQUIRE 
        \STATE List of candidate models $\mathbb{M}$ = \{$m_i$, ..., $m_n$\} \\
        \STATE Valid set COMET for each model $\mathbb{C}$ = \{$c_i$, ..., $c_n$\} \\
        \STATE Average Self-COMET for each model $\mathbb{S}$ = \{$s_i$, ..., $s_n$\} \\
        \STATE The number of models $n$ \\
        \STATE The number of ensemble models $e$ \\
		\ENSURE Selected Model Pool $\mathbb{P}$
	\end{algorithmic}
	\begin{algorithmic}[1]
		\FOR{$i \leftarrow 1 $ $to$ $n$}
        \STATE $weight$ = $\frac{(max(\mathbb{S})-min(\mathbb{S}))}{(max(\mathbb{C})-min(\mathbb{C}))}$
        \STATE $score_i$ = 
        \begin{small}
        $(c_i - min(\mathbb{C}))\cdot weight+(max(\mathbb{S}) - s_i)$
        \end{small}
        \ENDFOR
		\STATE Add the highest score model to candidates list $\mathbb{P}$ = \{  $m_{top}$ \} \\
        \WHILE {$|\mathbb{P}|$ \textless \ $e$ }
        \STATE $ index$ = 
        \begin{small}
        $\mathop{\arg\min}\limits_{i} \frac{1}{|\mathbb{M}-\mathbb{P}|}$
        $\sum\limits_{i \in {\mathbb{M}-\mathbb{P}},j\in{\mathbb{P}}} BLEU(i,j)$
        \end{small}
        \STATE Add $m_{index}$ to candidate list $\mathbb{P}$
        \ENDWHILE
		\STATE \textbf{return} $\mathbb{P}$
    \end{algorithmic}
\end{algorithm}

\subsubsection{Boosted Self-COMET-based Model Ensemble (BSCE)}
After we get plenty of fine-tuned models, how to search for the best combination for the ensemble model is a difficult question. Inspired by~\citet{zeng-etal-2021-wechat}, we propose a Boosted Self-COMET-based Ensemble (BSCE) algorithm, as shown in algorithm~\ref{alg:1}. Since the existing boosted Self-BLEU-based pruning strategy~\cite{zeng-etal-2021-wechat} is designed for achieving higher BLEU scores with high efficiency, it can not help obtain better COMET scores. Therefore, we adapt it to COMET scores. Then, we can obtain the best ensemble models from n top models by a greedy search strategy. 

The algorithm takes as input a list of n strong single models $\mathbb{M}$, COMET scores on the development set for each model $\rm \mathbb{C}$, average Self-COMET scores for each model $\rm \mathbb{S}$, the number of models $n$, and the expected number of ensemble models $e$. The algorithm returns a set $\rm \mathbb{P}$ consisting of $e$ selected models. We calculate the weighted score for each model (line 2). The weight (line 3) calculated is a trade-off between the development set COMET score and the Self-COMET score since the performance and the diversity play the same key role in ensemble~\cite{zeng-etal-2021-wechat}. Then the set $\rm \mathbb{P}$ initially contains the model $m_{top}$ has the highest weighted score. Next, we iteratively re-compute the average Self-COMET between the remaining models in `$\rm \mathbb{M}-\mathbb{P}$' and selected models in $\rm \mathbb{P}$, based on which we select the model that has a minimum Self-COMET score into $\rm \mathbb{P}$.

\section{Experiments and Results}

\subsection{Setting}
The implementation of our models is based on Fairseq\footnote{https://github.com/pytorch/fairseq}.
All the single models in the first pre-training stage are carried out on 8 NVIDIA V100 GPUs (32 GB memory of each). And all the models in the second fine-tuning stage are conducted on 4 NVIDIA V100 GPUs. We use the Adam optimizer with $\beta_{1}$ = 0.9, $\beta_{2}$ = 0.998. 
The batch size are set to 8192 and 4096 tokens per GPU for pre-training and fine-tuning, respectively. We set the ``update-freq'' parameter to 2 and 1 for both stages.
The learning rate is set to 0.0005 and 0.0004 for two stages, respectively. We use the warmup step to 4000.
We calculate COMET\footnote{https://unbabel.github.io/COMET/html/index.html} score for all experiments which is officially recommended.

English and German sentences are segmented by Moses\footnote{http://www.statmt.org/moses/}. We apply punctuation normalization and Truecasing. We use byte pair encoding BPE~\cite{sennrichetal2016bpe} with 32K operations. For the post-processing, we apply de-truecaseing and de-tokenizing on the English and German translations with the scripts provided in Moses.

\subsection{Dataset}
The data statistics of the two stages are shown in Table \ref{t:train_data}. For the general pre-training, the bilingual data is the combination of all parallel data in WMT21. For monolingual data, we use the News Crawl, Common Crawl, and Extended Common Crawl. For synthetic data generation, we back-translate all the target monolingual data and forward-translate the source monolingual data. For the in-domain fine-tuning, we use all the training, valid, and testing data of the wmt20 chat task as our training data. For monolingual data, we select the Taskmaster-1~\cite{byrne-etal-2019-taskmaster} corpus to build the pseudo-paired data using the method described in Section~\ref{speaker-aware}.

\begin{table}[!t]
\centering
\scalebox{0.70} {
\begin{tabular}{lrr}
\toprule
 & General pre-training & In-domain fine-tuning  \\
\midrule
Bilingual Data    & 74.8M & 17,847 \\
Source Mono Data  & 332.8M & 302,079 \\
Target Mono Data  & 237.9M & - \\
\bottomrule
\end{tabular}
}
\caption{Statistics of all training data.} 
\label{t:train_data}
\end{table}
 
\subsection{Results}

We report COMET scores~\cite{rei-etal-2020-comet} on the validation set (generally, beam size = 5 and length penalty = 0.6).

\paragraph{Pre-training and Fine-tuning.}

The results in Table~\ref{main_res} show that all pre-trained models outperform the baseline models trained on the chat training data. We observe that in-domain fine-tuning of the pre-trained models always gives large gains even on the in-domain pseudo data. We also find that the performance of different model architectures comes close after in-domain fine-tuning. Though these models perform similarly, as they have different architectures or are trained on different data, they generate diverse translations and show a cumulative effect when ensemble. 

\begin{table}[t]
\centering
\newcommand{\tabincell}[2]{\begin{tabular}{@{}#1@{}}#2\end{tabular}}
\scalebox{0.80}{
\setlength{\tabcolsep}{0.5mm}{
\begin{tabular}{lcc}
\toprule
\multirow{1}{*}{\textbf{Models}} & \multicolumn{1}{c}{$\textbf{En$\rightarrow$De}$} &  \multicolumn{1}{c}{$\textbf{De$\rightarrow$En}$} \\
\midrule
Chat baseline w/o context      &0.403  &0.588 \\
Chat baseline w context       &0.376  &0.680 \\\hline
Pre-trained deeper model w/o context   &0.544  &0.865 \\
+ in-domain genuine data w/ context (FT1) &0.772   &0.905 \\
+ in-domain pseudo data w/ context (FT2)    &{0.767} &{0.903} \\
+ in-domain both data w/ context (FT3)  &{0.781} &{0.908} \\\hline
Pre-trained wider model w/o context   &0.604  &0.879 \\
+ in-domain genuine data w/ context (FT4) &0.782   &0.908 \\
+ in-domain pseudo data w/ context (FT5)    &{0.779} &{0.906} \\
+ in-domain both data w/ context (FT6)  &\textbf{0.785} &\textbf{0.909} \\
\bottomrule
\end{tabular}}}
\caption{COMET scores on the Valid set for both pre-trained models, and each of fine-tuned on (i) in-domain genuine data, (ii) in-domain pseudo data, and  (iii) both in-domain data.}
\label{main_res} 
\end{table}
\label{ssec:cs}

\paragraph{Final Submissions.}
Table~\ref{submission} shows the results of our primary submission on both the validation and test set. Note that all candidate models with different architectures or trained with different data are used for the ensemble. We find that our BSCE is effective in both directions (more analyses are shown in Section~\ref{bsce}). Inspired by~\citet{wang-EtAl:2020:WMT1}, we also tried large beam size. Finally, our primary system achieves the highest results among all submissions\footnote{\url{https://wmt-chat-task.github.io/}}.

\begin{table}[t]
\centering
\newcommand{\tabincell}[2]{\begin{tabular}{@{}#1@{}}#2\end{tabular}}
\scalebox{0.80}{
\setlength{\tabcolsep}{3.5mm}{
\begin{tabular}{lcc}
\toprule
\multirow{1}{*}{\textbf{Models}} & \multicolumn{1}{c}{$\textbf{En$\rightarrow$De}$} &  \multicolumn{1}{c}{$\textbf{De$\rightarrow$En}$} \\
\midrule
Best Single Model      &{0.785} &{0.909} \\
+ Normal Ensemble      &{0.788} &{0.908} \\
+ BSCE                 &{0.790} &{0.911} \\
+ BSCE + Large beam (*)      &\textbf{0.792} &\textbf{0.913} \\\hline
Official results on the Test set  \\\hline
+ BSCE + Large beam (*)     &\textbf{0.810} &\textbf{0.946} \\
Best Official     &\textbf{0.810} &\textbf{0.946} \\
\bottomrule
\end{tabular}}}
\caption{Valid set COMET scores for ensemble with different strategies and the official COMET results of our submissions. `*' indicates the primary system of our submissions.}
\label{submission} 
\end{table}
\label{ssec:speaker}

\section{Analysis}
\subsection{Effect of Speaker Tags}
As shown in Table~\ref{speaker}, we observe that the performance in both directions improves with the addition of tags, which is consistent with~\citet{moghe-hardmeier-bawden:2020:WMT}. It shows that adding the speaker tag indeed can improve the chat translation performance.

\begin{table}[t]
\centering
\newcommand{\tabincell}[2]{\begin{tabular}{@{}#1@{}}#2\end{tabular}}
\scalebox{0.80}{
\setlength{\tabcolsep}{0.5mm}{
\begin{tabular}{lcc}
\toprule
\multirow{1}{*}{\textbf{Models}} & \multicolumn{1}{c}{$\textbf{En$\rightarrow$De}$} &  \multicolumn{1}{c}{$\textbf{De$\rightarrow$En}$} \\
\midrule
FT6 + no tag      &{0.779} &{0.904} \\
FT6 + speaker      &\textbf{0.785} &\textbf{0.909} \\
\bottomrule
\end{tabular}}}
\caption{Valid set COMET scores for fine-tuning with speaker tags .}
\label{speaker} 
\end{table}
\label{ssec:speaker}

\subsection{Effect of Prompt-based Context Modeling (PCM)}
As shown in Table~\ref{context}, we investigate the effect of the context. The bilingual context involves the utterance in mixed language. Therefore, we investigate the different contexts with prompt learning. The results show that the models achieve slight performance gains with suitable context. And using context in the same language was more beneficial than the mixed context, which is consistent with previous work~\cite{moghe-hardmeier-bawden:2020:WMT}.

\begin{table}[t]
\centering
\newcommand{\tabincell}[2]{\begin{tabular}{@{}#1@{}}#2\end{tabular}}
\scalebox{0.80}{
\setlength{\tabcolsep}{0.5mm}{
\begin{tabular}{lcc}
\toprule
\multirow{1}{*}{\textbf{Models}} & \multicolumn{1}{c}{$\textbf{En$\rightarrow$De}$} &  \multicolumn{1}{c}{$\textbf{De$\rightarrow$En}$} \\
\midrule
FT6 + w/o context     &\textbf{0.782} &\textbf{0.905} \\\hline
using previous context (mix language)  \\\hline
FT6 + w/ PCM (+ 1 prev)      &{0.781} &\textbf{0.905} \\
FT6 + w/ PCM (+ 2 prev)      &{0.779} &{0.901} \\
FT6 + w/ PCM (+ 3 prev)      &{0.775} &{0.897} \\\hline
using previous context (same language)  \\\hline
FT6 + w/ PCM (+ 1 prev)      &\textbf{0.785} &\textbf{0.909} \\
FT6 + w/ PCM (+ 2 prev)      &{0.784} &\textbf{0.909} \\
FT6 + w/ PCM (+ 3 prev)      &{0.782} &{0.904} \\
\bottomrule
\end{tabular}}}
\caption{Valid set COMET scores for fine-tuning with different contexts. The numbers before ``prev'' indicate the number of preceding utterances used as context.}
\label{context} 
\end{table}
\label{ssec:cs}

\subsection{Effect of Boosted Self-COMET-based Ensemble (BSCE)}
\label{bsce}
Inspired by the boosted Self-BLEU-based ensemble~\cite{zeng-etal-2021-wechat}, we propose the Boosted Self-COMET-based Ensemble. To verify its superiority, we first select the top 10 models with different architecture and training data. The results are shown in the ``+Normal Ensemble'' of Table~\ref{submission}. For the BSCE, we need to get the translation result of every model to calculate the Self-COMET. After that, we only need to perform the inference process once. Then, we can select the best models for the ensemble. Here, we select 10 models and 4 models for $\textbf{En$\rightarrow$De}$ and $\textbf{De$\rightarrow$En}$, respectively. The results are shown in ``+BSCE'' of Table~\ref{submission}. Based on it, we obtain better results after using the large beam (beam sizes of 9 and 8 for $\textbf{En$\rightarrow$De}$ and $\textbf{De$\rightarrow$En}$, respectively). These results show the effectiveness of our BSCE method.

\section{Conclusions}
We investigate the pre-training-then-fine-tuning paradigm to build chat translation systems, which are some effective transformer-based architectures. Our systems are also built on several popular data augmentation methods such as back-translation, forward-translation, and knowledge distillation. In the fine-tuning, we enhance our system by speaker-aware in-domain data generation, speaker adaptation, prompt-based context modeling, target denoising fine-tuning~\cite{meng-etal-2020-wechat}, and boosted self-COMET-based model ensemble. Our systems achieve 0.810 and 0.946 COMET~\cite{rei-etal-2020-comet} scores on English$\rightarrow$German and German$\rightarrow$English, respectively. These COMET scores are the highest among all submissions.
\section*{Acknowledgements}
The research work described in this paper has been supported by the National Key R\&D Program of China (2020AAA0108005) and the National Nature Science Foundation of China (No. 61976015, 61976016, 61876198 and 61370130). The authors would like to thank the anonymous reviewers for their valuable comments and suggestions to improve this paper.
\bibliography{anthology,custom}

\begin{thebibliography}{30}
\expandafter\ifx\csname natexlab\endcsname\relax\def\natexlab#1{#1}\fi

\bibitem[{Buschbeck et~al.(2022)Buschbeck, Mell, Exel, and
  Huck}]{buschbeck2022hi}
Bianka Buschbeck, Jennifer Mell, Miriam Exel, and Matthias Huck. 2022.
\newblock “hi, how can i help you?” improving machine translation of
  conversational content in a business context.
\newblock In \emph{Proceedings of the 23rd Annual Conference of the European
  Association for Machine Translation}, pages 189--198.

\bibitem[{Byrne et~al.(2019)Byrne, Krishnamoorthi, Sankar, Neelakantan,
  Goodrich, Duckworth, Yavuz, Dubey, Kim, and
  Cedilnik}]{byrne-etal-2019-taskmaster}
Bill Byrne, Karthik Krishnamoorthi, Chinnadhurai Sankar, Arvind Neelakantan,
  Ben Goodrich, Daniel Duckworth, Semih Yavuz, Amit Dubey, Kyu-Young Kim, and
  Andy Cedilnik. 2019.
\newblock \href {https://doi.org/10.18653/v1/D19-1459} {Taskmaster-1: Toward a
  realistic and diverse dialog dataset}.
\newblock In \emph{Proceedings of EMNLP-IJCNLP}, pages 4516--4525.

\bibitem[{Edunov et~al.(2018)Edunov, Ott, Auli, and
  Grangier}]{edunov-etal-2018-understanding}
Sergey Edunov, Myle Ott, Michael Auli, and David Grangier. 2018.
\newblock \href {https://doi.org/10.18653/v1/D18-1045} {Understanding
  back-translation at scale}.
\newblock In \emph{Proceedings of EMNLP}, pages 489--500, Brussels, Belgium.

\bibitem[{Farajian et~al.(2020)Farajian, Lopes, Martins, Maruf, and
  Haffari}]{farajian-etal-2020-findings}
M.~Amin Farajian, Ant{\'o}nio~V. Lopes, Andr{\'e} F.~T. Martins, Sameen Maruf,
  and Gholamreza Haffari. 2020.
\newblock \href {https://www.aclweb.org/anthology/2020.wmt-1.3} {Findings of
  the {WMT} 2020 shared task on chat translation}.
\newblock In \emph{Proceedings of WMT}, pages 65--75.

\bibitem[{Gain et~al.(2022)Gain, Appicharla, Ekbal, Chelliah, Chennabasavraj,
  and Garera}]{gain2022low}
Baban Gain, Ramakrishna Appicharla, Asif Ekbal, Muthusamy Chelliah, Soumya
  Chennabasavraj, and Nikesh Garera. 2022.
\newblock Low resource chat translation: A benchmark for hindi--english
  language pair.
\newblock In \emph{Proceedings of the 15th biennial conference of the
  Association for Machine Translation in the Americas (Volume 1: Research
  Track)}, pages 83--96.

\bibitem[{Gain et~al.(2021)Gain, Haque, and Ekbal}]{gain2021not}
Baban Gain, Rejwanul Haque, and Asif Ekbal. 2021.
\newblock Not all contexts are important: The impact of effective context in
  conversational neural machine translation.
\newblock In \emph{2021 International Joint Conference on Neural Networks
  (IJCNN)}, pages 1--8. IEEE.

\bibitem[{Hoang et~al.(2018)Hoang, Koehn, Haffari, and
  Cohn}]{hoang2018iterative}
Vu~Cong~Duy Hoang, Philipp Koehn, Gholamreza Haffari, and Trevor Cohn. 2018.
\newblock \href {https://doi.org/10.18653/v1/W18-2703} {Iterative
  back-translation for neural machine translation}.
\newblock In \emph{Proceedings of WMT}, pages 18--24, Melbourne, Australia.

\bibitem[{Kim and Rush(2016)}]{kim-rush-2016-sequence}
Yoon Kim and Alexander~M. Rush. 2016.
\newblock \href {https://doi.org/10.18653/v1/D16-1139} {Sequence-level
  knowledge distillation}.
\newblock In \emph{Proceedings of EMNLP}, pages 1317--1327, Austin, Texas.

\bibitem[{Liang et~al.(2021{\natexlab{a}})Liang, Meng, Chen, Xu, and
  Zhou}]{liang-etal-2021-modeling}
Yunlong Liang, Fandong Meng, Yufeng Chen, Jinan Xu, and Jie Zhou.
  2021{\natexlab{a}}.
\newblock \href {https://doi.org/10.18653/v1/2021.acl-long.444} {Modeling
  bilingual conversational characteristics for neural chat translation}.
\newblock In \emph{Proceedings of ACL}, pages 5711--5724.

\bibitem[{Liang et~al.(2022{\natexlab{a}})Liang, Meng, Xu, Chen, and
  Zhou}]{liang2022msctd}
Yunlong Liang, Fandong Meng, Jinan Xu, Yufeng Chen, and Jie Zhou.
  2022{\natexlab{a}}.
\newblock \href {https://doi.org/10.18653/v1/2022.acl-long.186} {{MSCTD}: A
  multimodal sentiment chat translation dataset}.
\newblock In \emph{Proceedings of ACL}, pages 2601--2613, Dublin, Ireland.

\bibitem[{Liang et~al.(2022{\natexlab{b}})Liang, Meng, Xu, Chen, and
  Zhou}]{liang-etal-2022-scheduled}
Yunlong Liang, Fandong Meng, Jinan Xu, Yufeng Chen, and Jie Zhou.
  2022{\natexlab{b}}.
\newblock \href {https://doi.org/10.18653/v1/2022.acl-long.300} {Scheduled
  multi-task learning for neural chat translation}.
\newblock In \emph{Proceedings of ACL}, pages 4375--4388, Dublin, Ireland.

\bibitem[{Liang et~al.(2021{\natexlab{b}})Liang, Meng, Zhang, Chen, Xu, and
  Zhou}]{liang2021infusing}
Yunlong Liang, Fandong Meng, Ying Zhang, Yufeng Chen, Jinan Xu, and Jie Zhou.
  2021{\natexlab{b}}.
\newblock Infusing multi-source knowledge with heterogeneous graph neural
  network for emotional conversation generation.
\newblock In \emph{Proceedings of AAAI}, volume~35, pages 13343--13352.

\bibitem[{Liang et~al.(2022{\natexlab{c}})Liang, Meng, Zhang, Chen, Xu, and
  Zhou}]{liang2022emotional}
Yunlong Liang, Fandong Meng, Ying Zhang, Yufeng Chen, Jinan Xu, and Jie Zhou.
  2022{\natexlab{c}}.
\newblock Emotional conversation generation with heterogeneous graph neural
  network.
\newblock \emph{Artificial Intelligence}, 308:103714.

\bibitem[{Liang et~al.(2021{\natexlab{c}})Liang, Zhou, Meng, Xu, Chen, Su, and
  Zhou}]{liang-etal-2021-towards}
Yunlong Liang, Chulun Zhou, Fandong Meng, Jinan Xu, Yufeng Chen, Jinsong Su,
  and Jie Zhou. 2021{\natexlab{c}}.
\newblock \href {https://aclanthology.org/2021.emnlp-main.6} {Towards making
  the most of dialogue characteristics for neural chat translation}.
\newblock In \emph{Proceedings of EMNLP}, pages 67--79.

\bibitem[{Liu et~al.(2021)Liu, Sun, and Wang}]{liu2021recent}
Siyou Liu, Yuqi Sun, and Longyue Wang. 2021.
\newblock Recent advances in dialogue machine translation.
\newblock \emph{Information}, 12(11):484.

\bibitem[{Ma et~al.(2020)Ma, Zhang, and Zhou}]{ma-etal-2020-simple}
Shuming Ma, Dongdong Zhang, and Ming Zhou. 2020.
\newblock \href {https://doi.org/10.18653/v1/2020.acl-main.321} {A simple and
  effective unified encoder for document-level machine translation}.
\newblock In \emph{Proceedings of ACL}, pages 3505--3511.

\bibitem[{Meng et~al.(2020)Meng, Yan, Liu, Gao, Zeng, Zeng, Li, Chen, Zhou,
  Liu, and Zhou}]{meng-etal-2020-wechat}
Fandong Meng, Jianhao Yan, Yijin Liu, Yuan Gao, Xianfeng Zeng, Qinsong Zeng,
  Peng Li, Ming Chen, Jie Zhou, Sifan Liu, and Hao Zhou. 2020.
\newblock \href {https://aclanthology.org/2020.wmt-1.24} {{W}e{C}hat neural
  machine translation systems for {WMT}20}.
\newblock In \emph{Proceedings of WMT}, pages 239--247, Online.

\bibitem[{Meng and Zhang(2019)}]{meng2019dtmt}
Fandong Meng and Jinchao Zhang. 2019.
\newblock \href {https://doi.org/10.1609/aaai.v33i01.3301224} {{DTMT:} {A}
  novel deep transition architecture for neural machine translation}.
\newblock In \emph{Proceedings of AAAI}, pages 224--231.

\bibitem[{Moghe et~al.(2020)Moghe, Hardmeier, and
  Bawden}]{moghe-hardmeier-bawden:2020:WMT}
Nikita Moghe, Christian Hardmeier, and Rachel Bawden. 2020.
\newblock \href {https://www.aclweb.org/anthology/2020.wmt-1.58} {The
  university of edinburgh-uppsala university's submission to the wmt 2020 chat
  translation task}.
\newblock In \emph{Proceedings of WMT}, pages 471--476.

\bibitem[{Ranzato et~al.(2016)Ranzato, Chopra, Auli, and
  Zaremba}]{ranzato2015exposurebias}
Marc'Aurelio Ranzato, Sumit Chopra, Michael Auli, and Wojciech Zaremba. 2016.
\newblock \href {http://arxiv.org/abs/1511.06732} {Sequence level training with
  recurrent neural networks}.
\newblock In \emph{Proceedings of ICLR}.

\bibitem[{Rei et~al.(2020)Rei, Stewart, Farinha, and
  Lavie}]{rei-etal-2020-comet}
Ricardo Rei, Craig Stewart, Ana~C Farinha, and Alon Lavie. 2020.
\newblock \href {https://doi.org/10.18653/v1/2020.emnlp-main.213} {{COMET}: A
  neural framework for {MT} evaluation}.
\newblock In \emph{Proceedings of EMNLP}, pages 2685--2702, Online.

\bibitem[{Sennrich et~al.(2016)Sennrich, Haddow, and
  Birch}]{sennrichetal2016bpe}
Rico Sennrich, Barry Haddow, and Alexandra Birch. 2016.
\newblock \href {https://doi.org/10.18653/v1/P16-1162} {Neural machine
  translation of rare words with subword units}.
\newblock In \emph{Proceedings of ACL}, pages 1715--1725, Berlin, Germany.

\bibitem[{Shazeer et~al.(2020)Shazeer, Lan, Cheng, Ding, and
  Hou}]{https://doi.org/10.48550/arxiv.2003.02436}
Noam Shazeer, Zhenzhong Lan, Youlong Cheng, Nan Ding, and Le~Hou. 2020.
\newblock \href {https://doi.org/10.48550/ARXIV.2003.02436} {Talking-heads
  attention}.

\bibitem[{Vaswani et~al.(2017)Vaswani, Shazeer, Parmar, Uszkoreit, Jones,
  Gomez, Kaiser, and Polosukhin}]{vaswani2017attention}
Ashish Vaswani, Noam Shazeer, Niki Parmar, Jakob Uszkoreit, Llion Jones,
  Aidan~N Gomez, \L~ukasz Kaiser, and Illia Polosukhin. 2017.
\newblock \href
  {https://proceedings.neurips.cc/paper/2017/file/3f5ee243547dee91fbd053c1c4a845aa-Paper.pdf}
  {Attention is all you need}.
\newblock In \emph{Proceedings of NIPS}, pages 5998--6008.

\bibitem[{Wang et~al.(2021)Wang, Yan, Meng, and Zhou}]{wang2021selective}
Fusheng Wang, Jianhao Yan, Fandong Meng, and Jie Zhou. 2021.
\newblock \href {https://doi.org/10.18653/v1/2021.acl-long.504} {Selective
  knowledge distillation for neural machine translation}.
\newblock In \emph{Proceedings of ACL}, pages 6456--6466, Online.

\bibitem[{Wang et~al.(2020)Wang, Tu, Wang, Ding, Ding, and
  Shi}]{wang-EtAl:2020:WMT1}
Longyue Wang, Zhaopeng Tu, Xing Wang, Li~Ding, Liang Ding, and Shuming Shi.
  2020.
\newblock \href {https://www.aclweb.org/anthology/2020.wmt-1.60} {Tencent ai
  lab machine translation systems for wmt20 chat translation task}.
\newblock In \emph{Proceedings of WMT}, pages 481--489.

\bibitem[{Yan et~al.(2020)Yan, Meng, and Zhou}]{yanetal2020multi}
Jianhao Yan, Fandong Meng, and Jie Zhou. 2020.
\newblock \href {https://doi.org/10.18653/v1/2020.emnlp-main.77} {Multi-unit
  transformers for neural machine translation}.
\newblock In \emph{Proceedings of EMNLP}, pages 1047--1059, Online.

\bibitem[{Zeng et~al.(2021)Zeng, Liu, Li, Ran, Meng, Li, Xu, and
  Zhou}]{zeng-etal-2021-wechat}
Xianfeng Zeng, Yijin Liu, Ernan Li, Qiu Ran, Fandong Meng, Peng Li, Jinan Xu,
  and Jie Zhou. 2021.
\newblock \href {https://aclanthology.org/2021.wmt-1.23} {{W}e{C}hat neural
  machine translation systems for {WMT}21}.
\newblock In \emph{Proceedings of WMT}, pages 243--254, Online. Association for
  Computational Linguistics.

\bibitem[{Zhang et~al.(2018)Zhang, Xiong, and
  Su}]{zhang-etal-2018-accelerating}
Biao Zhang, Deyi Xiong, and Jinsong Su. 2018.
\newblock \href {https://doi.org/10.18653/v1/P18-1166} {Accelerating neural
  transformer via an average attention network}.
\newblock In \emph{Proceedings of ACL}, pages 1789--1798, Melbourne, Australia.
  Association for Computational Linguistics.

\bibitem[{Zhang et~al.(2019)Zhang, Feng, Meng, You, and
  Liu}]{zhang-etal-2019-bridging}
Wen Zhang, Yang Feng, Fandong Meng, Di~You, and Qun Liu. 2019.
\newblock \href {https://doi.org/10.18653/v1/P19-1426} {Bridging the gap
  between training and inference for neural machine translation}.
\newblock In \emph{Proceedings of ACL}, pages 4334--4343, Florence, Italy.

\end{thebibliography}
\bibliographystyle{acl_natbib}




\end{document}